# TourBERT: A pretrained language model for the tourism industry.

Veronika Arefieva[1], Roman Egger[2]

[1]Johannes Keppler University of Linz

[2]Salzburg University of Applied Sciences

[1]{veroare@me.com }

[2]{roman.egger@fh-salzburg.ac.at}

**Introduction**

The Bidirectional Encoder Representations from Transformers (BERT) is currently the most important and state-of-the-art natural language model (Tenney et al., 2019) since its launch in 2018 by Google. BERT Large, which is based on a Transformer architecture, is considered one of the most powerful language models with 24 layers, 16 attention heads, and 340 million parameters (Lan et al. 2019). BERT is a pretrained model and can be fine-tuned to perform numerous downstream tasks such as text classification, question answering, sentiment analysis, extractive summarization, named entity recognition, or sentence similarity (Egger, 2022). The model was pretrained on a huge English corpus in a self-supervised way. Raw texts from a BookCorpus of over 11,000 books and English Wikipedia were used to generate this state-of-the-art model. From this, it can already be concluded that BERT was trained on a huge generic corpus (Edwards et al., 2020) . However, it has been shown in the past that for domain-specific applications and downstream tasks, it is helpful to pretrain BERT on a large domain-specific corpus to enable it to learn the linguistic peculiarities better (Gururangan et al., 2020). For example, BERT variants have been pretrained for the financial sector (FinBERT) (Araci, 2019), the medical sector (Clinical BERT) (Alsentzer et al., 2019), for biomedical texts (BioBERT) (Lee et al., 2020), or SciBERT (Beltagy and Cohan, 2019) for biomedical and computer science.

Tourism is one of the most important economic sectors in the world (Hollenhorst et al., 2014), and its services have many characteristics that distinguish them from other products. Services are not tangible and cannot be tested in advance, which is why the customer assumes an increased risk before starting the trip. The service is co-created together with the customer, so the customer is an active co-creator of the service. Services are subject to the uno-actu principle, which means they are produced at the same time as they are consumed, and they are considered bilateral, i.e. a reciprocal relationship between persons (Chehimi, 2014). In addition, tourism services are relatively expensive compared to everyday products and have an intercultural dimension. All this means that tourism services are extremely description-intensive (Dooolin et al., 2002). In addition to detailed descriptions by the supply side, user-generated content is becoming increasingly important (Yu and Egger, 2021). Whether on review platforms such as TripAdvisor or social media channels such as Twitter, Facebook or Instagram, people everywhere are sharing their travel experiences, thus influencing other users (Daxböck et al, 2021). This content is of particular importance for tourism providers, as they are losing control over UGC (Saraiva, 2013).

The automated analysis of texts using natural language processing methods is therefore becoming increasingly important for both academia and the tourism industry (Egger and Gokce, 2022).

In order to meet the requirements of tourism, we introduce TourBERT in this paper. It was pretrained on 3.6 million tourist reviews and about 50k descriptions of tourist services and attractions and sights from more than 20 different countries around the world. The intercultural context, in particular, leads to linguistic peculiarities that BERT as a general language model cannot cope with. In the following, we introduce TourBERT and describe how it was trained and evaluated.

We therefore pretrained TourBERT from scratch with 1M steps using BERT-Base architecture with WordPiece tokenizer and our crawled, tourism specific vocabulary with the same vocabulary size as BERT-Base. For the pretraining procedure we followed official BERT repo recommendations.

**Technical description**

TourBERT has BERT-Base-uncased as an underlying architecture and was trained from scratch. So no initial checkpoints were used, like it was done for BioBERT or FinBERT. The whole corpus was pre-processed by lowercasing the data, splitting them into sentences using punctuation as separators. A custom WordPice tokenizer was trained to create a custom input for the TourBERT model using 30.522 tokens in total which is the same number as BERT-Base has.

The pretraining was performed for 1M steps on a single TPU instance provided by Google Colab Pro which took about three days in total.

**TourBERT Model evaluation**

In order to evaluate TourBERT, both quantitative and qualitative measurements were applied. A supervised task, namely sentiment classification, was used for the quantitative evaluation.

**Evaluation Task 1: Sentiment Classification**

To perform classification using BERT a number of different options are available. First, a softmax layer on top of the BERT architecture could be used, which is one of the most widely used approaches. Second, an LSTM neural network can be used as a separate classification model. This approach is useful if an input example cannot be represented as a single vector, i.e. the length of an input text is significantly greater than the maximum input length allowed by the BERT-model. In our case, we decided, however, to use a single feed-forward layer on top of the BERT architecture, as this is a simple way to construct a classifier, and this approach is widely used to benchmark different BERT models against each other. After a single feed-forward layer is attached on top of the BERT architecture, all layers of the BERT model itself are frozen, i.e. only the classifier is trained.

The authors are aware that this does not usually yield state-of-the-art results, but the goal of this evaluation was not to achieve the highest score on a given dataset but to show that the quality of TourBERT embeddings surpasses the BERT-Base model.

The sentiment task was performed on two different datasets. First, on a Tripadvisor hotel review dataset from Ray et al. (2021) which is a available at: https://ieee-dataport.org/documents/hotel-reviews-around-world-sentiment-values-and-review-ratings-different-categories

This dataset has three Labels: {-1: "negative", 0:"neutral", 1: "positive"} and a total of 69.308 reviews.

The second dataset is the "515k reviews from Europe hotels" dataset available at https://www.kaggle.com/jiashenliu/515k-hotel-reviews-data-in-europe; we used only reviews which have either negative or positive labels and thus turned this problem into a binary classification with the two following labels:

{-1: "negative", 1: "positive"}. We sampled 35.000 positive and 35.000 negative reviews resulting in 70.000 samples in total.

Table 1 shows the evaluation results for both datasets and that TourBERT outperforms BERT-Base in that tasks.

|  | Dataset 1 (multi-label) | | Dataset 2 (binary classification) | |
|---|---|---|---|---|
|  | CrossEntropy Loss | Accuracy | CrossEntropy Loss | Accuracy |
| BERT-Base | 0.4527 | 0.8178 | 0.2341 | 0.9230 |
| **TOURBert** | **0.3168** | **0.8638** | **0.1515** | **0.9540** |

**Table 1.** Results of supervised evaluation for TourBERT and BERT-Base on two sentiment classification datasets.

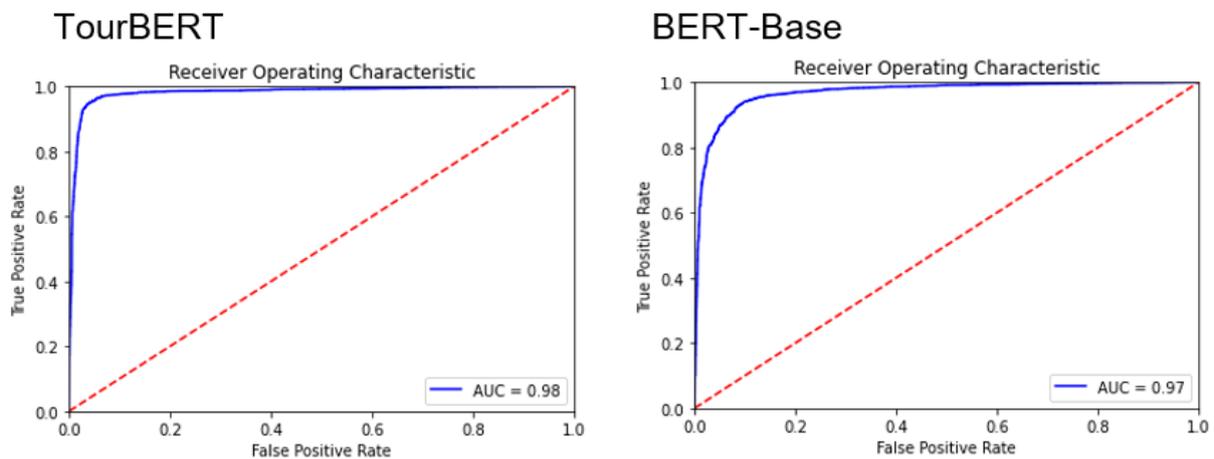

**Figure 1.** The Area under ROC-Curve (AUC) scores for TourBERT and BERT-Base

For the qualitative evaluation, unsupervised tasks and a user study was performed. On the one hand, a topic modeling task, synonyms search, and a within-vocabulary words similarity distribution task were designed.

**Evaluation Task 2: Unsupervised Evaluation / Visualization of Photos**

The first unsupervised evaluation was the visualization of photos in the Tensorboard Projector. Therefore a dataset of 48 photos showing different tourism activities like sports activities, visiting sights, shopping, and others were used. Next, a sample of 622 people was engaged to perform manual labeling of these photos by assigning two bi-gram tags to each foto. These annotations were then visualized using the TensorBoard projector API, which allows visualizing original photos on a 2D- or a 3D-plot centered at their respective cluster centers. Finally, the evaluation was done after performing UMAP by inspecting and comparing the group's separation quality on the plot.

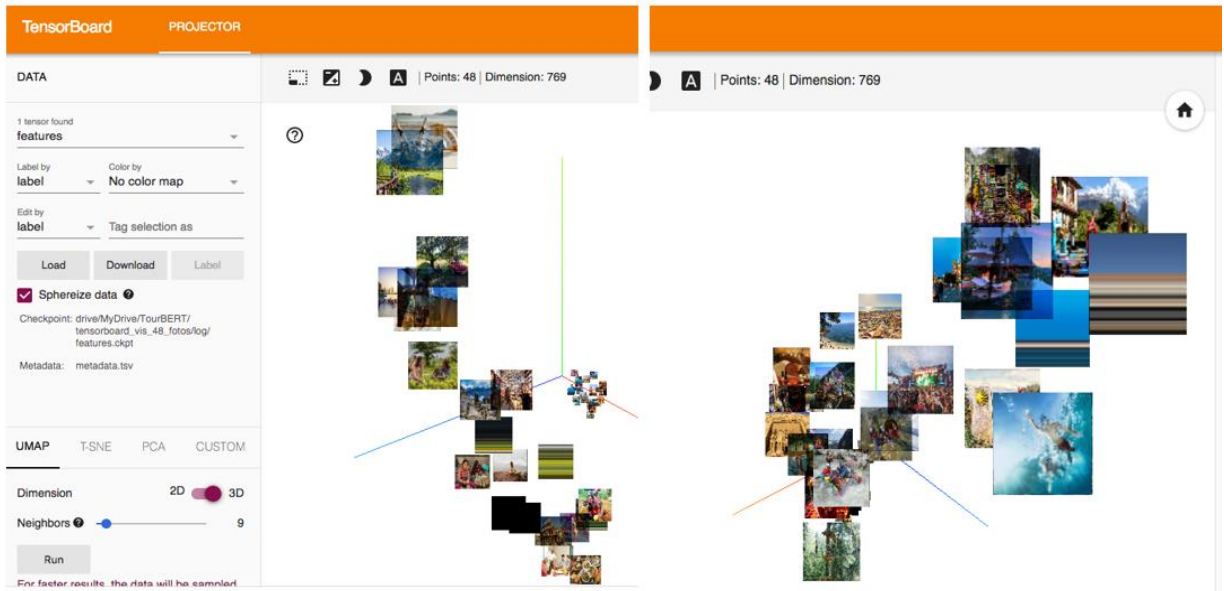

**Figure 2.** BERT-Base / TensorBoard Projector

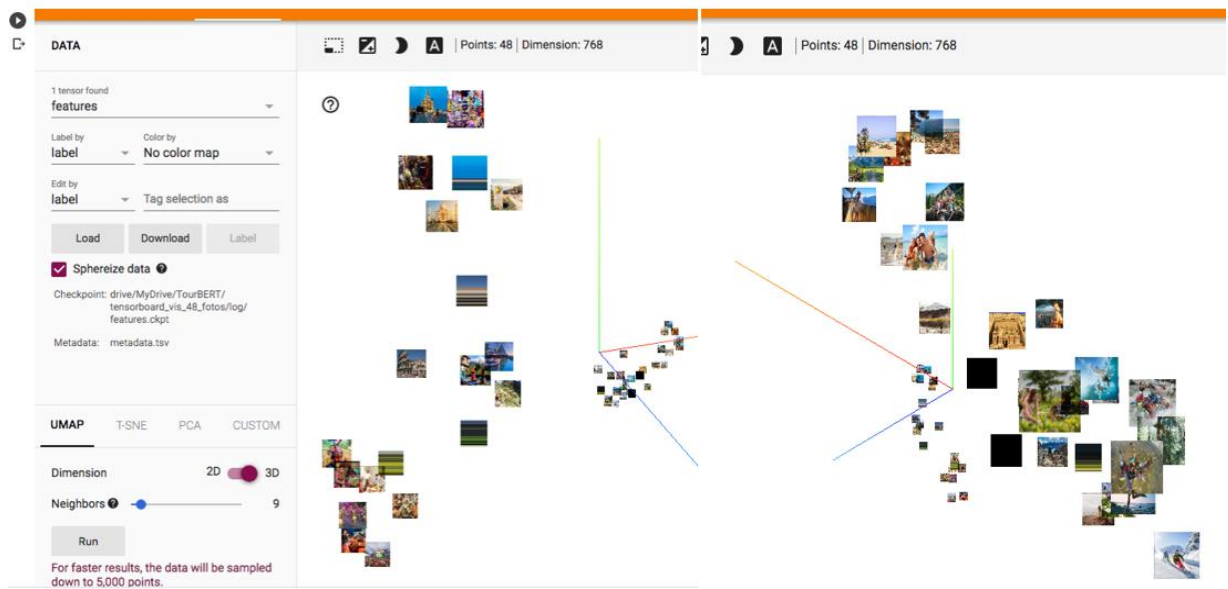

**Figure 3**. TourBERT / TensorBoard Projector

The purpose of such a visualization is to evaluate the separation of clusters which naturally result from the down-projection method. We can see that TourBERT vectors result in better separated groups and pictures within the same group have similar contents. By looking at the results produced using BERT-Base vectors, it can be observed that photos are heavily mixed and do not allow to identify well-separated groups.

**Evaluation Task 3: Unsupervised Evaluation / Topic Modeling**

A subsequent and unsupervised evaluation was done by applying a topic modeling approach. For this, 5000 Instagram posts from public accounts with the hashtag #wanderlust were crawled using the Python ScraPy library. Instagram is based on photos which are the primary source of information, while the textual description of Instagram posts is often either limited to hashtags, emojis and unrelated to the photo or completely missing. Therefore images were annotated using Google Cloud Vision API, and a TouBERT vector was generated for each photo annotation. Then a K-means clustering approach was used to group the annotation vectors. K was chosen using the silhouette score, which resulted in 25 clusters. Then a PCA down-projection method was used to visualize the points on a plot which allowed us to determine how distinct topic words are and whether they overlap. This approach did not aim to develop a new topic modeling approach as there are embedding-based approaches with a similar logic available like Top2Vec (Angelov, 2020) or BERTopic (Grootendorst, 2020), but only to compare the performance of BERT-Base vs. TourBERT. For the evaluation, an interactive visualization board using Python's Sklearn and Altair libraries was developed, similar to pyLDAVis for LDA topic modeling, containing two blocks. The block on the lift visualizes cluster centers on a 2D plot, where the size of a cluster center is proportional to the cluster population size. Each point is clickable and defines the output on the second block on the right. This is a horizontal bar chart, visualizing the top 15 most frequent words for a topic. The darker bars represent the word distribution within the entire dataset. The lighter bars show the word distribution within the selected topic. Evaluation can now be done on two aspects: the goodness of the topic separation (i.e. cluster centers do not overlap and are placed far away from each other) and how similar are words within the same cluster.

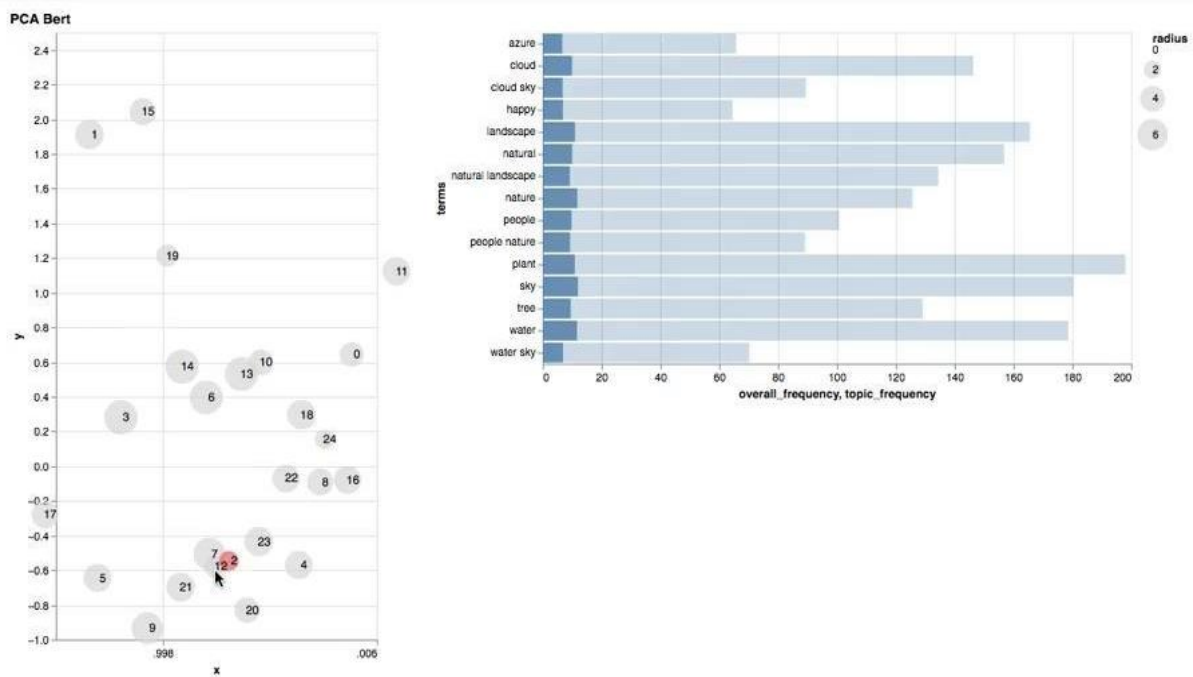

**Figure 4.** Topic Modeling – Results for BERT-Base

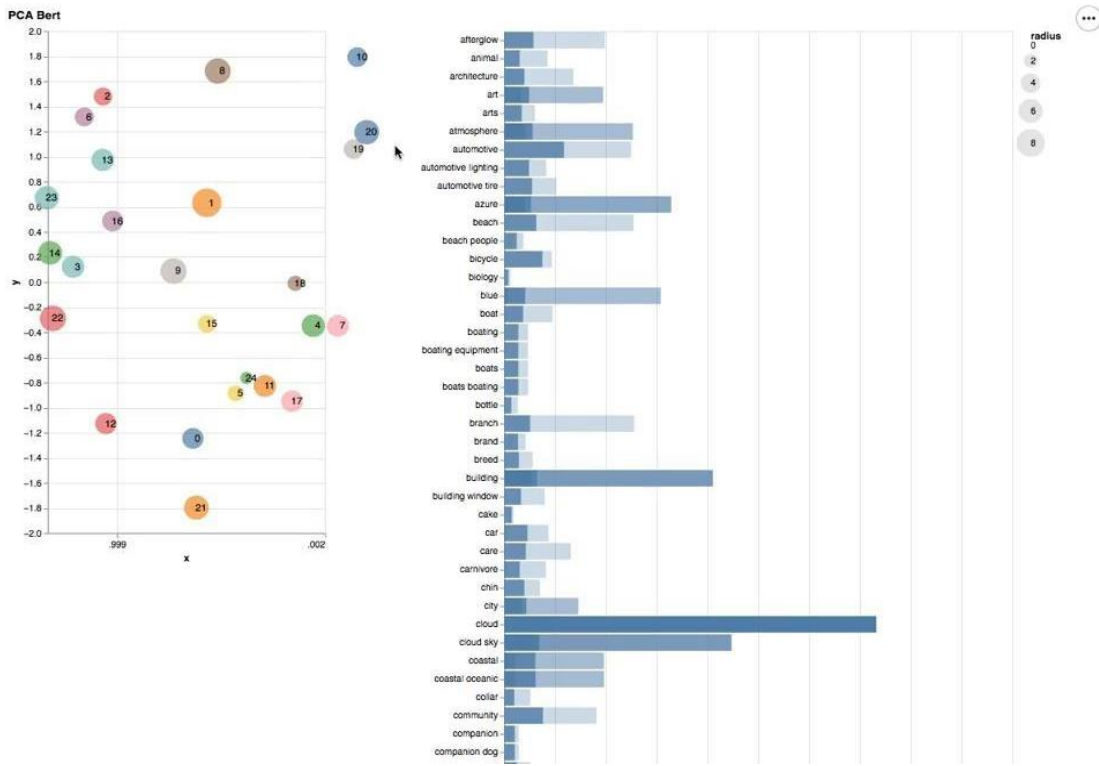

**Figure 5.** Topic Modeling – Results for TourBERT

From figures 4 and 5, we can see that cluster centers produced using down-projected TourBERT vectors are clearly better separated compared to those produced with BERT-Base ones. Some of the topic words can be seen in tables 2 and 3 for BERT-Base and TourBERT, respectively:

| | Topic Words |
|---|---|
| 0 | fashion, sleeve, shoulder, flash, flash photography, photography, street, street fashion, smile, hair, neck, eyewear, eyebrow, happy, sky |
| 1 | shades, tints, tints shades, plant, black, sky, shirt, bicycle, photography, white, font, sleeve, wood, building, automotive |
| 2 | sky, nature, water, landscape, plant, natural, cloud, people, tree, people nature, natural landscape, water sky, happy, cloud sky, azure |
| 3 | automotive, vehicle, sky, tire, font, plant, landscape, design, wood, art, building, rectangle, cloud, water, lighting |
| 4 | plant, natural, water, landscape, natural landscape, sky, ecoregion, cloud, tree, mountain, nature, cloud sky, highland, community, plant community |
| 5 | water, landforms, sky, coastal, coastal oceanic, oceanic, oceanic landforms, landscape, cloud, natural, beach, water sky, natural landscape, azure, plant |
| 6 | people, nature, sky, smile, people nature, sunglasses, flash, flash photography, photography, water, sleeve, care, vision, vision care, eyewear |
| 7 | landscape, sky, plant, cloud, natural, natural landscape, water, tree, building, nature, cloud sky, mountain, vehicle, people, blue |
| 8 | water, sky, cloud, landscape, plant, natural, natural landscape, resources, water resources, building, tree, mountain, cloud sky, water sky, nature |
| 9 | landscape, plant, natural, sky, water, natural landscape, nature, cloud, tree, grass, people, people nature, cloud sky, sky plant, wood |
| 10 | fashion, happy, sky, people, nature, photography, flash, flash photography, eyewear, smile, people nature, care, vision, vision care, plant |
| 11 | plant, sky, water, natural, landscape, ecoregion, tree, natural landscape, cloud, photography, fashion, flash, flash photography, smile, happy |
| 12 | plant, natural, landscape, water, natural landscape, sky, tree, dog, nature, grass, cloud, terrestrial, wood, people, landforms |
| 13 | building, sky, plant, window, vehicle, facade, tree, wood, design, house, automotive, tire, cloud, road, city |
| 14 | vehicle, automotive, sky, building, plant, tire, font, design, art, window, cloud, tree, wood, rectangle, lighting |
| 15 | plant, shades, tints, tints shades, sky, wood, black, fashion, bicycle, photography, rectangle, people, white, building, font |
| 16 | plant, water, natural, sky, landscape, natural landscape, cloud, ecoregion, mountain, tree, cloud sky, community, plant community, resources, water resources |
| 17 | landscape, plant, water, sky, natural, natural landscape, shades, tints, tints shades, tree, cloud, landforms, wood, coastal, coastal oceanic |
| 18 | fashion, sleeve, flash, flash photography, photography, street, street fashion, lip, shoulder, eyelash, eyebrow, smile, hairstyle, sky, neck |
| 19 | water, sky, equipment, cloud, equipment supplies, supplies, boating, boating equipment, boats, boats boating, landforms, boat, watercraft, coastal, coastal oceanic |
| 20 | water, landscape, natural, plant, sky, cloud, natural landscape, mountain, tree, nature, cloud sky, azure, highland, resources, water resources |
| 21 | plant, water, sky, nature, landscape, natural, cloud, tree, people, natural landscape, people nature, grass, cloud sky, mountain, building |
| 22 | sky, plant, cloud, water, landscape, building, natural, tree, natural landscape, mountain, cloud sky, window, nature, travel, road |
| 23 | plant, natural, sky, landscape, water, natural landscape, tree, cloud, nature, terrestrial, terrestrial plant, flower, grass, petal, wood |
| 24 | food, sky, cuisine, ingredient, recipe, tableware, dish, food tableware, ingredient recipe, water, tableware ingredient, staple, staple food, plate, produce |

**Table 2.** Topic words for 25 topics produced with BERT-Base vectors.

| | Topic Words |
|---|---|
| 0 | plant, sky, tree, building, road, landscape, wood, cloud, road surface, surface, grass, window, sky plant, leisure, water |
| 1 | diving, underwater, water, fluid, marine, equipment, biology, marine biology, organism, fish, water underwater, liquid, diving equipment, underwater diving, blue |
| 2 | beach, people, water, sky, people beach, cloud, nature, people nature, water sky, azure, happy, travel, beach people, coastal, coastal oceanic |
| 3 | landscape, mountain, natural, sky, cloud, natural landscape, plant, slope, tree, cloud sky, highland, snow, sky mountain, terrain, sky plant |
| 4 | font, art, arts, event, rectangle, brand, design, pattern, graphics, photography, happy, painting, magenta, logo, visual |
| 5 | building, sky, window, facade, tower, design, urban, city, cloud, urban design, plant, sky building, road, house, building window |
| 6 | water, sky, afterglow, cloud, dusk, atmosphere, landscape, natural, natural landscape, sky atmosphere, cloud sky, sunlight, sunset, water sky, tree |
| 7 | tableware, drinkware, table, bottle, cup, dishware, food, glass, wood, plant, furniture, device, stemware, kitchen, wine |
| 8 | people, nature, sky, people nature, flash, flash photography, photography, happy, water, smile, plant, cloud, leg, gesture, tree |
| 9 | water, sky, equipment, boat, watercraft, cloud, vehicle, lake, supplies, boating, boating equipment, boats, boats boating, equipment supplies, water sky |
| 10 | care, vision, vision care, sunglasses, sleeve, eyewear, goggles, glasses, sky, dress, fashion, smile, shirt, flash, flash photography |
| 11 | automotive, vehicle, tire, bicycle, wheel, motor, motor vehicle, automotive tire, vehicle automotive, sky, lighting, automotive lighting, car, plant, tire wheel |
| 12 | plant, landscape, natural, natural landscape, sky, tree, nature, grass, community, plant community, cloud, people, people nature, water, sky plant |
| 13 | sky, water, cloud, landscape, natural, atmosphere, cloud sky, blue, natural landscape, azure, plant, nature, tree, horizon, sunlight |
| 14 | water, natural, landscape, sky, natural landscape, cloud, plant, nature, mountain, resources, water resources, ecoregion, tree, cloud sky, water sky |
| 15 | temple, sky, building, architecture, plant, facade, city, cloud, art, travel, tree, leisure, sculpture, world, monument |
| 16 | nature, plant, people nature, people, sky, happy, tree, landscape, cloud, natural, water, grass, natural landscape, travel, leisure |
| 17 | wood, design, building, rectangle, interior, interior design, window, shades, tints, tints shades, property, font, furniture, flooring, plant |
| 18 | food, cuisine, ingredient, tableware, recipe, dish, food tableware, ingredient recipe, produce, staple, staple food, cuisine dish, tableware ingredient, plate, cake |
| 19 | fashion, street, street fashion, sleeve, eyewear, flash, flash photography, photography, shirt, happy, waist, smile, dress, design, shoe |
| 20 | lip, eyebrow, eyelash, smile, hair, chin, shoulder, skin, nose, forehead, hairstyle, neck, eye, lip chin, facial |
| 21 | plant, flower, tree, terrestrial, twig, landscape, terrestrial plant, natural, petal, natural landscape, branch, grass, wood, sky, flowering |
| 22 | water, natural, plant, landscape, landforms, natural landscape, fluvial, fluvial landforms, landforms streams, streams, resources, water resources, sky, watercourse, water water |
| 23 | water, landscape, landforms, natural, sky, coastal, coastal oceanic, oceanic, oceanic landforms, cloud, natural landscape, water sky, azure, resources, water resources |
| 24 | dog, plant, animal, carnivore, breed, dog breed, fawn, sky, terrestrial, working, working animal, companion, companion dog, collar, grass |

**Table 3.** Topic words for 25 topics produced with TourBERT vectors.

To have a better understanding of the topics' quality, we output the top-10 nearest samples for each cluster and look at the photos which the samples were produced for like shown in figures 6 and 7 below. Each figure contains a table with the first column showing words for a given topic and all subsequent columns showing top-10 most similar samples, i.e. photos for that topic.

**Figure 6.** The first six topics with cluster words and top-10 most similar images, produced by the K-Means model using TourBERT vectors.

**Figure 7.** The first six topics with cluster words and top-10 most similar images, produced by the K-Means model using BERT-Base vectors.

Comparing both models' results, we can see that clusters achieved with TourBERT vectors are much more homogenous within the clusters and heterogenous between the clusters compared to those for BERT-Base, which sometimes include relatively dissimilar photos belonging to the same topic, like in topic 3.

To further investigate the quality of each topic produced by the model and prove our assumptions, we conducted a user study to statistically evaluate the results, which is described in detail in the next section.

**Evaluation Task 4: Unsupervised Evaluation / User Study**

For the same dataset of images and annotations, a user study was designed. Therefore a set of the 10 most similar photos for each of the 25 clusters of BERT-Base and TourBERT was created, and users were asked to evaluate how similar the photos within each of the 50 clusters are on a 7-point Likert scale with possible answers ranging from "very similar" to "very different". This evaluation approach allows for getting an intersubjective perception of the cluster qualities, similar to measuring the intercoder reliability in qualitative studies. The image clusters were shown to the participants in a rotating manner, i.e. randomly alternating.

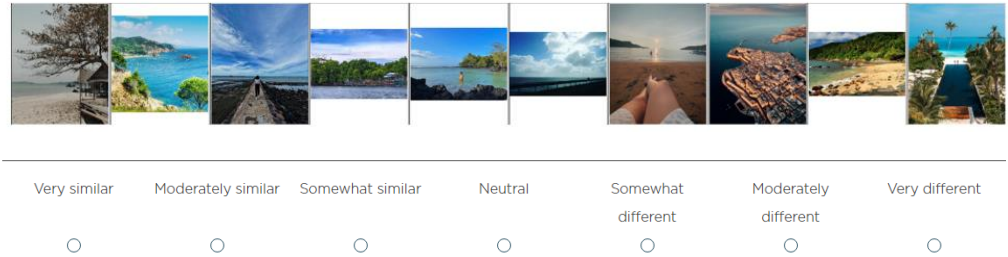

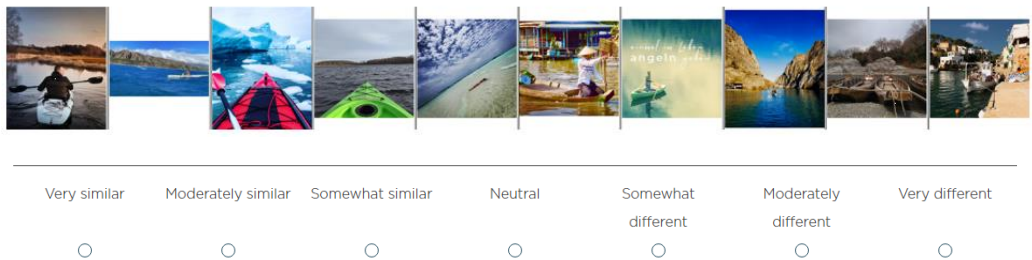

**Figure 6.** Two examples of image clusters

For the evaluation of study results a pairwise t-test was performed with SPSS. The coding ranged from [1 – very similar] to [7 – very different], and the mean values were 3,75 and 2,5 for BERT-Base and TourBERT respectively at a highly significant level (Sig. two-sided ,000). The effect size was measured with Cohen´s d and yielded with 0,517 a medium level effect.

**Paired Samples Statistics**

|  |  | Mean | N | Std. Deviation | Std. Error Mean |
|---|---|---|---|---|---|
| Pair 1 | BERT | 3,7759 | 82 | ,71655 | ,07913 |
|  | TourBERT | 2,5239 | 82 | ,61724 | ,06816 |

**Paired Samples Statistics**

|  |  | Mean | N | Std. Deviation | Std. Error Mean |
|---|---|---|---|---|---|
| Pair 1 | BERT | 3,7759 | 82 | ,71655 | ,07913 |
|  | TourBERT | 2,5239 | 82 | ,61724 | ,06816 |

**Paired Samples Test**

| | | Paired Differences | | | | | | | |
|---|---|---|---|---|---|---|---|---|---|
| | | Mean | Std. Deviation | Std. Error Mean | 95% Confidence Interval of the Difference | | t | df | Sig. (2-tailed) |
| | | | | | Lower | Upper | | | |
| Pair 1 | BERT - TourBERT | 1,25201 | ,51773 | ,05717 | 1,13825 | 1,36577 | 21,898 | 81 | ,000 |

**Paired Samples Effect Sizes**

| | | | Standardizer[a] | Point Estimate | 95% Confidence Interval | |
|---|---|---|---|---|---|---|
| | | | | | Lower | Upper |
| Pair 1 | BERT - TourBERT | Cohen's d | ,51773 | 2,418 | 1,986 | 2,846 |
| | | Hedges' correction | ,52015 | 2,407 | 1,977 | 2,833 |

a. The denominator used in estimating the effect sizes.

Cohen's d uses the sample standard deviation of the mean difference.

Hedges' correction uses the sample standard deviation of the mean difference, plus a correction factor.

**Table 4.** Results of the paired samples mean comparison

These results show that the similarity of the annotated images was perceived significantly better with TourBERT than with BERT-Base.

**Evaluation Task 5: Synonyms Search**

According to the assumption that BERT-Base, due to its general trained corpus, achieves more generic results than the TourBERT model trained on a tourism-specific corpus, it was assumed that a similarity search of tourism-related terms would turn out better with TourBERT than with BERT-Base. For similarity search, we choose words displayed in the first row of the table, which are: "authenticity", "experience", "entrance", and so on. We output the top-8 most similar words for each word, which can be seen in the tables below.

| authenticity | experience | entrance | attraction | ticket | destination | guide | transfer | sightseeing | service |
|---|---|---|---|---|---|---|---|---|---|
| legitimacy | teach | shelter | attractions | tickets | dying | companion | recovery | trees | vessel |
| sincerity | heal | entrances | restaurant | fare | choice | entry | exchange | fireworks | authority |
| competence | communicate | archway | hotel | fares | lame | visit | imaging | shops | headquarters |
| authorship | consume | gate | exhibit | card | address | database | restoring | pacing | facility |
| flexibility | learn | roof | pavilion | trains | exit | forum | sale | comedy | workshop |
| integrity | eat | causeway | nightclub | bus | partner | workshop | comparison | prostitutes | circulation |
| conscience | consider | tenants | mall | metro | correction | access | recovering | sidewalk | companion |
| characterization | experiences | exit | ballroom | freight | priorities | google | screening | nights | operation |

**Figure 7.** Synonyms Search with BERT-Base

| authenticity | experience | entrance | attraction | ticket | destination | guide | transfer | sightseeing | service |
|---|---|---|---|---|---|---|---|---|---|
| uniqueness | experince | entry | destination | tickets | spot | ##guide | transfers | exploring | sevice |
| ambiance | expereince | enterance | feature | entry | attraction | guides | transport | sights | services |
| originality | experiance | admittance | landmark | entrance | place | tourguide | pickup | attractions | staff |
| intimacy | adventure | admission | place | wristband | point | guid | transportation | exploration | personnel |
| charm | experiences | ticket | institution | admission | itinerary | driver | journey | nightlife | hospitality |
| accuracy | enjoyment | fee | musuem | fee | hotspot | interpreter | limousine | hiking | personel |
| flare | opportunity | carpark | spot | pass | venture | guiding | shuttle | outings | frontdesk |
| warmth | expere | payment | site | tix | hangout | narrator | pickups | excursions | housekeeping |

**Figure 8.** Synonyms Search with TourBERT

When comparing synonyms produced by BERT-Base and TourBERT, one can see that TourBERT almost perfectly captures a tourism-specific meaning of the word. On the contrary, BERT-Base captures a more generic meaning of the same words. For example, TourBERT associates the word "destination" with words like "spot" and "attraction" and place, whereas BERT-Base considers the same word "destination" to be similar with terms like "dying", "choice", and "lame".

**Summary and model usage**

All evaluation tasks have proven the performance of TourBERT for use in the tourism-specific context. TourBERT outperforms BERT-Base in all tasks and thus represents a suitable language model for both academia and the tourism industry. The Tensorflow checkpoint of TourBERT has been converted to the PyTorch binary format and is released on Hugging Face Hub at the following URL: https://huggingface.co/veroman/TourBERT
One can simply load the TourBERT model

and the tokenizer using the following three
lines of code:

```python
from transformers import AutoTokenizer, AutoModel

model_auto = AutoModel.from_pretrained('veroman/TourBERT')

tokenizer = AutoTokenizer.from_pretrained("veroman/TourBERT",
                                          do_lowercase=True,
                                          truncation=True, padding="max_length",
                                          model_max_length=128)
```

TourBERT Release V1: 16.01.2022

## References:

Alsentzer, E., Murphy, J. R., Boag, W., Weng, W. H., Jin, D., Naumann, T., & McDermott, M. (2019). Publicly available clinical BERT embeddings. *arXiv preprint arXiv:1904.03323*.

Angelov, D. (2020). Top2vec: Distributed representations of topics. *arXiv preprint arXiv:2008.09470.*

Araci, D. (2019). Finbert: Financial sentiment analysis with pre-trained language models. *arXiv preprint arXiv:1908.10063*.

Beltagy, I., Lo, K., & Cohan, A. (2019). Scibert: A pretrained language model for scientific text. *arXiv preprint arXiv:1903.10676*.

Chehimi, N. (2014). Tourist Information Search. In *The Social Web in the Hotel Industry* (pp. 49-70). Springer Gabler, Wiesbaden.

Daxböck, J., Dulbecco, M. L., Kursite, S., Nilsen, T. K., Rus, A. D., Yu, J., & Egger, R. (2021). The Implicit and Explicit Motivations of Tourist Behaviour in Sharing Travel Photographs on Instagram: A Path and Cluster Analysis. In *Information and Communication Technologies in Tourism 2021* (pp. 244-255). Springer, Cham.

Doolin, B., Burgess, L., & Cooper, J. (2002). Evaluating the use of the Web for tourism marketing: a case study from New Zealand. *Tourism management*, *23*(5), 557-561.

Edwards, A., Camacho-Collados, J., De Ribaupierre, H., & Preece, A. (2020, December). Go simple and pre-train on domain-specific corpora: On the role of training data for text classification. In *Proceedings of the 28th International Conference on Computational Linguistics* (pp. 5522-5529).

Egger, R. (2022) Text Representations and Word Embeddings. Vectorizing Textual Data. In: Egger, R. (Ed.) *Applied Data Science in Tourism. Interdisciplinary Aprochaches, Methodologies and Applicaitions.* Springer (forthcoming).

Egger, R. and Gokce, E. (2022) Natural Language Processing: An Introduction. In: Egger, R. (Ed.) *Applied Data Science in Tourism. Interdisciplinary Aprochaches, Methodologies and Applicaitions.* Springer (forthcoming).

Grootendorst, M. (2020) *BERTopic: Leveraging BERT and c-TF-IDF to create easily interpretable topics*. Available at: https://maartengr.github.io/BERTopic/index.html

Gururangan, S., Marasović, A., Swayamdipta, S., Lo, K., Beltagy, I., Downey, D., & Smith, N. A. (2020). Don't stop pretraining: adapt language models to domains and tasks. *arXiv preprint arXiv:2004.10964.*

Hollenhorst, S. J., Houge-Mackenzie, S., & Ostergren, D. M. (2014). The trouble with tourism. *Tourism Recreation Research*, *39*(3), 305-319.

Lan, Z., Chen, M., Goodman, S., Gimpel, K., Sharma, P., & Soricut, R. (2019). Albert: A lite bert for self-supervised learning of language representations. *arXiv preprint arXiv:1909.11942*.

Lee, J., Yoon, W., Kim, S., Kim, D., Kim, S., So, C. H., & Kang, J. (2020). BioBERT: a pre-trained biomedical language representation model for biomedical text mining. *Bioinformatics*, *36*(4), 1234-1240.

Saraiva, J. P. D. P. M. (2013). *Web 2.0 in restaurants: insights regarding TripAdvisor's use in Lisbon* (Doctoral dissertation).

Tenney, Ian, Dipanjan Das, and Ellie Pavlick. "BERT rediscovers the classical NLP pipeline." *arXiv preprint arXiv:1905.05950* (2019).

Yu, J., & Egger, R. (2021). Tourist Experiences at Overcrowded Attractions: A Text Analytics Approach. In *Information and Communication Technologies in Tourism 2021* (pp. 231-243). Springer, Cham.